\pdfoutput=1 
\documentclass[letterpaper]{article} 
\usepackage[preprint]{aaai2027}  
\usepackage[hyphens]{url}  
\usepackage{graphicx} 
\usepackage{natbib}  
\usepackage{caption} 
\usepackage{algorithm}
\usepackage{algorithmic}

\usepackage{newfloat}
\usepackage{listings}
\DeclareCaptionStyle{ruled}{labelfont=normalfont,labelsep=colon,strut=off} 
\floatstyle{ruled}
\newfloat{listing}{tb}{lst}{}
\floatname{listing}{Listing}

\usepackage{booktabs}

\usepackage{makecell}          
\usepackage{amssymb}           
\newcommand{\yes}{\textcolor{green!60!black}{\checkmark}}
\newcommand{\no}{\textcolor{red!80!black}{\texttimes}}
\usepackage{comment}
\usepackage{pifont}

\usepackage{multirow}
\usepackage{tabularx}
\usepackage{colortbl}
\usepackage{xcolor}
\usepackage{adjustbox}
\newcolumntype{C}{>{\centering\arraybackslash}X}
\title{Bootstrapping Niche Multilingual Code Translation \\ via Reinforcement Learning with Execution-Based Verifiable Supervision}

\author{
    Kouki Yuki\textsuperscript{\rm 1},
    Jie Zeng\textsuperscript{\rm 2},
    Kyoko Ogawa\textsuperscript{\rm 3},
    Ryunosuke Ikeda\textsuperscript{\rm 4},
    Yohei Kobashi\textsuperscript{\rm 5},\\
    Takeshi Kojima\textsuperscript{\rm 5},
    Ikuya Yamada\textsuperscript{\rm 6,\rm 7},
    Yusuke Iwasawa\textsuperscript{\rm 5},
    Yutaka Matsuo\textsuperscript{\rm 5}
}
\affiliations{
    \mbox{\textsuperscript{\rm 1}National Institute of Technology, Numazu College, Japan}\quad
    \mbox{\textsuperscript{\rm 2}Seikei University, Japan}\quad
    \mbox{\textsuperscript{\rm 3}Osaka Metropolitan University, Japan}\quad
    \mbox{\textsuperscript{\rm 4}Recruit Co., Ltd., Japan}\quad
    \mbox{\textsuperscript{\rm 5}The University of Tokyo, Japan}\quad
    \mbox{\textsuperscript{\rm 6}Tokyo University of Science, Japan}\quad
    \mbox{\textsuperscript{\rm 7}Studio Ousia, Japan}\\
    t.kojima@weblab.t.u-tokyo.ac.jp
}

\begin{document}

\maketitle

\begin{abstract}
Code translation must preserve executable behavior across many programming languages, yet neural code translation has largely focused on a few popular languages such as C++, Java, and Python. This leaves a niche, many-to-many setting where parallel supervision is sparse, producing plausible but non-executable translations.
We address this setting with preference-based reinforcement learning driven by execution-based supervision.
Our pipeline firstly expands verifiable seed Python programs into a multilingual pool of execution-validated codes.
Using the pool, A base LLM generates translation candidates across language pairs, which we label by their execution outcomes.
The resulting preferences are used train a reward model that scores cross-language translation quality.
Finally, we optimize our base LLMs with GRPO over 600 directed language pairs (25 × 24) using the reward model as a signal. 
To evaluate the niche translation capability, we introduce HumanEval-X++, an execution-based benchmark that extends HumanEval-X to a broad many-to-many language space. 
We evaluate our approach using Qwen-3.5 4B and 9B models. On HumanEval-X++ and existing benchmarks, it yields consistent gains over the untrained baselines. In particular, the 4B model achieves an average improvement of 13\% across all languages on HumanEval-X++, with a gain of 21\% on mid-tier languages.
Our study establishes a reliable approach of data generation, training, and benchmarking, paving the way toward further bootstrapping the quality of many-to-many translation for programming languages.
\end{abstract}


\section{Introduction}

\begin{table*}[t]
\centering
\small
\setlength{\tabcolsep}{5pt}
\begin{tabular}{@{}lcccc@{}}
\toprule
\multirow{3}{*}{\makecell[l]{Model}} & \multicolumn{2}{c}{\makecell{Code Translation}} & \multirow{3}{*}{\makecell{Utilize Execution-Based\\Verification}} & \multirow{3}{*}{\makecell{Training Method}} \\
\cmidrule(lr){2-3}
 & \makecell{\# of Source} & \makecell{\# of Target} & & \\
\midrule
TransCoder \citep{transcoder} & 3 & 3 & \no & Back-Translation \\
TransCoder-ST \citep{transcoder-st} & 3 & 3 & \yes & Self-Training \\
CoTran \citep{cotran} & 2 & 2 & \yes & SFT + RL (PPO) \\
CodeGeeX \citep{codegeex} & 7 & 7 & \no & SFT \\
ExeCoder \citep{execoder} & 3 & 3 & \no & SFT \\
OORL \citep{oorl} & 2 & 6 & \yes & RL (GEPO + REINFORCE++) \\
EffiReasonTrans \citep{effireasontrans} & 3 & 3 & \yes & SFT + RL (GRPO) \\
BootTrans \citep{boottrans} & 3 & 3 & \yes & RL (GRPO) \\
CodePivot \citep{li2026codepivot} & 1 & 9 & \yes & SFT + RL (GRPO) \\
\textbf{Ours} & \textbf{25} & \textbf{25} & \yes & \textbf{RL (GRPO) w/ Learned Reward Model} \\
\bottomrule
\end{tabular}
\caption{Related work of bootstrapping many-to-many code translation by model training. Language counts (`\# of Source', `\# of Target') refer to the languages covered by each method's training pipeline. 
`Utilize Execution-Based Verification' indicates whether actual unit-test execution is used to verify and reward translations during training.
}
\label{tab:related_work_main}
\end{table*}

Modern software is written in many programming languages with different syntax, idioms, and type systems. Code translation is an important capability for software migration, modernization, reuse, and interoperability, but practical translation models must work reliably across diverse language pairs while preserving executable behavior. 
In many-to-many code translation setting, supervision is sparse and language-specific knowledge is less reliable, increasing the risk that plausible-looking translations fail to preserve executable behavior \citep{transcoder-st}.
 
Prior work has improved code translation capability of Large Language Models (LLMs) through execution-validated self-training, feedback-based optimization, and reasoning-augmented training \citep{transcoder,transcoder-st,cotran,effireasontrans}. 
However, these efforts mainly target a small number of popular language pairs or individual translation directions such as C++, Java, and Python \citep{transcoder}, leaving less-represented languages underexplored despite practical needs (\textcolor{black}{\ding{226} Section \ref{sec:related_work}}).
This leaves open how to construct scalable execution-derived supervision for bootstrapping niche, many-to-many translation based on reinforcement learning (RL). 
Many-to-many translation often requires us to treat a large portion of low-resource languages, whose supervised translation data is sparse as the target language becomes niche and long-tail.
 
We address this gap by introducing \textit{NicheCodeTranslator}, which applies  RL to long-tail many-to-many code translation, using reward model-based supervision constructed through a scalable, verification-driven bootstrapping pipeline (\textcolor{black}{\ding{226} Section \ref{sec:nichecodetranslator}}).
Specifically, the pipeline firstly expands verifiable seed Python programs into a multilingual pool of execution-validated codes.
Using the pool, A base LLM generates translation candidates across language pairs, which we label success of failure by their execution outcomes.
The resulting preferences are used to train a reward model that scores cross-language translation quality.
Finally, we optimize our base LLMs with GRPO over 600 directed language pairs (25 × 24) using the reward model as a signal.
 
We also propose HumanEval-X++, a new execution-based benchmark for long-tail many-to-many code translation to enable controlled evaluation of translation quality across popular and less-represented directions (\textcolor{black}{\ding{226} Section \ref{sec:humanevalxpp}}). Building on HumanEval-X \citep{codegeex}, HumanEval-X++ extends the original number of evaluation pair for code translation (source: 6 languages -> target: 6 languages) to a broader many-to-many language space (source: 6 languages -> target: 25 languages) by curating target-language prompts, executable function signatures, and test suites with comparable problem semantics across languages. 

We evaluate our approach on HumanEval-X++ and CodeScope \citep{yan-etal-2024-codescope}, an existing code translation benchmark, using the Qwen-3.5 4B and 9B models, which serve as strong baselines for code translation at these parameter scales (\textcolor{black}{\ding{226} Section \ref{sec:experiment}}).
Our experiments show consistent gains over the non-trained baselines.
In particular, the 4B model achieves an average improvement of 13\% across all languages on HumanEval-X++, with a 21\% gain on mid-tier languages, indicating that the improvements are especially pronounced for niche languages relative to popular ones.

 

\begin{figure*}[t]
  \centering
  \includegraphics[width=0.99\textwidth]{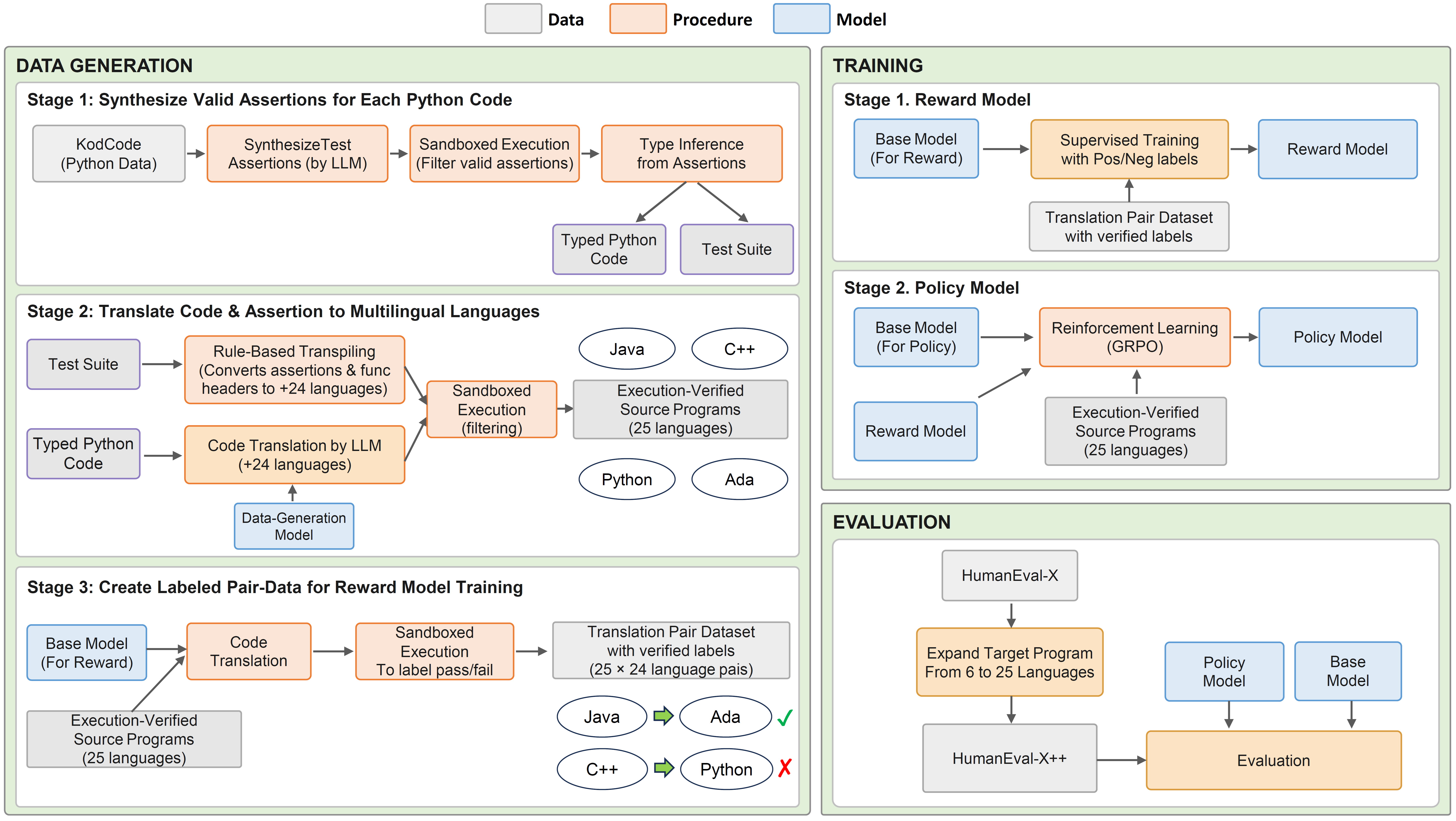}
  \caption{Overview of our pipeline. \textbf{Data Generation:}
Stage1 synthesizes assertions for each KodCode Python seed with sandboxed execution-based filtering, followed by type inference.
Stage2 exports the code and assertion pairs to 24 further languages. Only candidates passing all transpiled tests are admitted, yielding execution-verified source programs in 25 languages.
Stage3 builds the labeled pair data by translating each source into the other 24 languages, and sandboxed execution labels every candidate positive or negative, covering $25 \times 24 = 600$ directions.
\textbf{Training:}
Stage1 trains a reward model on these execution-verified labels.
A single model scores all translation directions.
Stage2 optimizes the policy with GRPO over the verified source pool.
The learned reward replaces per-language sandboxes at rollout time: RL stably and efficiently scales to 600 directions.
\textbf{Evaluation:}
HumanEval-X is extended to HumanEval-X++ by the same rule-based transpilation, broadening the target from 6 to 25 languages.
The models are evaluated by executing the generated code against the transpiled test suites.
}
  \label{fig:overview}
\end{figure*}

\section{Related Work}
\label{sec:related_work}

Code translation has long been used in a range of settings, including the migration of legacy assets, the integration of systems composed of different languages, and specification-change scenarios such as updating Python 2 to Python 3 or migrating COBOL to Java. Unlike natural-language translation, code translation prioritizes functional equivalence over surface-level similarity. For this reason, execution-based evaluation—grounded in compilability, executability, and unit-test pass rate—has been emphasized over static metrics such as BLEU. In the many-to-many setting in particular, the same functionality must be reproduced across multiple languages without failure.
 
TransCoder \citep{transcoder}, a pioneering work in this field, achieved unsupervised many-to-many translation (e.g., Denoising Auto-Encoding, Back Translation, etc) among C++, Java, and Python, demonstrating that translation is feasible without parallel data. 
However, its training does not exploit code-executability information (e.g., compilation or test results), and its focusing languages are limited to three.
The consecutive line of work, TransCoder-ST \citep{transcoder-st} has proposed self-training which train the model using self-generated translation samples that passes unit-test.
CoTran \citep{cotran} is an early attempt to bring execution feedback directly into the training loop. It fine-tunes models on Java-Python translation by interleaving supervised fine-tuning with PPO-based reinforcement learning using the signal. However, its coverage remains limited to two languages.
 
CodeGeeX \citep{codegeex} is a large-scale model pre-trained on 23 individual languages (not translation pairs) with roughly 850 billion tokens, followed by supervised fine-tuning on translation pairs constituting 7 languages.
Additionally, it introduced HumanEval-X, a multilingual extension of HumanEval. This made it possible to compare multilingual code tasks, including translation, under a common test specification, contributing significantly to research progress and benchmark development.
ExeCoder \citep{execoder} progressively incorporates execution-related representations such as function semantics, abstract syntax trees (AST), and variable dependencies (DFG), improving accuracy in translation among C++, Java, and Python. 
Nevertheless, its target languages remain limited to three.
 
OORL \citep{oorl} introduces on-policy reinforcement learning (REINFORCE++) with rewards derived from unit tests. It additionally employs preference optimization via GEPO based on the equivalence of intermediate representations. However, it performs neither automatic test synthesis nor training-data augmentation.
EffiReasonTrans \citep{effireasontrans} augments code translation by distilling (source code, reasoning, target code) triplets from a stronger reasoning model, retaining only samples that survive automated syntax validation and functional testing.
It then applies supervised fine-tuning followed by GRPO with a dual reward combining unit-test pass rate and an output-length tolerance term. 
BootTrans \citep{boottrans} takes a resource-rich pivot language (Python) with abundant code--test pairs, ports the test oracles to the target languages via rule-based templates, expanding translation matrix and unlocks reverse and cross-lingual directions absent from the seed data.
However, training is confined to three languages.
CodePivot \citep{li2026codepivot} trains only Python-to-others directions (9 targets) via SFT and GRPO by utilizing an execution reward, relying on zero-shot transfer for the remaining directions among 10 languages.

While such prior work has demonstrated that RL with execution-based verification is the reliable training signal for bootstrapping code translation, it needs to presupposes a compiler, runtime, and test harness for every language during the rollout for RL.
This requirement hinders the practical feasibility when expanding the number of target languages covering niche and long-tail ones at training, e.g., time-consuming verification and unexpected runtime error during the unit-test.
Our work addresses this limitation by decoupling the reward from per-language execution infrastructure. 
We train a reward model that predicts functional unit-test result, and optimize the policy with GRPO using this learned reward. 
This allows the training pipeline to stably scale to 25 target languages, an order of magnitude beyond prior work, while remaining grounded in functional correctness.
As a reference, prior work \citep{zhu2024deepseek} reports that reward-model-based approaches outperform test-based ones in coding (but not translation) tasks.

\section{NicheCodeTranslator}
\label{sec:nichecodetranslator}

\subsection{Generating Verified Translation Data}

Translation quality is defined by functional equivalence \cite{transcoder}; its practical proxy is a unit-test suite executable in the \emph{target} language. Training data must therefore pair each translation with such tests. Because candidate code in any language can be sampled cheaply from LLMs, the scarce half of each pair is the tests. 
Our pipeline resolves this scarcity by \emph{propagating the ability to verify} from Python to every other language, under three rules: (1)~every artifact is derived from, and validated against, a Python reference solution drawn from large public corpora \cite{kodcode}; (2)~tests and interfaces cross language boundaries only through deterministic, rule-based transpilation \cite{multipl-e}, never through a generative model; 
and (3)~candidates enter the dataset only by passing sandboxed execution, never by fallible LLM judgment. 

The pipeline proceeds in three stages (Figure~\ref{fig:overview}): Stage~1 synthesizes valid assertions from seed Python codes,
Stage~2 exports it to the other 24 languages, and Stage~3 generates translation data in every language for reward model training.

\subsubsection{Stage 1: Synthesizing Valid Assertions}
\label{sec:data:stage1}

\paragraph{Seed curation.}
We selected KodCode \cite{kodcode} as the source of Python seed code because it provides a large collection of verified Python programs synthesized from multiple sources and spanning a diverse range of programming tasks and domains.
Every instance in KodCode is a self-contained, function-level task consisting of a problem statement, a reference solution, and unit tests.
For the curation, we retain instances that satisfy the following conditions: (1) the solution defines exactly one function and no classes, (2)
it imports only standard-library modules, and (3)
the target function is actually invoked in attached assertion tests.

\paragraph{Test synthesis.}
A Data-generation LLM produces candidate tests, restricted to single assertions over literals so that rule-based transpilers can convert them deterministically, and only assertions that pass sandboxed execution against the reference solution are kept. Each surviving assertion thus records a verified input-output pair, which we exploit next.

\paragraph{Type annotation from tests.}
Stage~2 will transpile each Python signature into a function header for every target language, and statically typed targets require concrete parameter and return types, which KodCode solutions lack. Rather than ask an LLM to guess them, we recover these types by AST analysis of the assertion literals, emitting one annotated variant per observed type combination. 
Because every assertion has survived execution, each recovered type is backed by confirmed behavior of the reference solution.
During type annotation, a single instance may yield multiple variants if more than one type combination is consistent with  assertion.

\paragraph{Test validation.}
Tests are regenerated for the typed code and filtered by the same sandboxed execution, and examples with few assertions are removed. 
The surviving seeds constitute the canonical form with cross-lingual verifiability: typed functions with execution-verified literal assertions.

\subsubsection{Stage 2: Translating Code \& Assertions to 24 Languages}
\label{sec:data:stage2}

For each seed, rule-based transpilers \cite{multipl-e} convert the typed Python signature into a target-language function header and translates each assertion into a target-language test independently. Untranslatable items are conservatively dropped. Because this channel is deterministic, a translated test is exactly as trustworthy as its Python original.
Verified source programs are then produced for every language. The data-generation LLM is given the problem statement and the typed reference solution and asked to translate it, with the transpiled header imposed as a fixed prefix of the generated code. 
This guarantees by construction that the generated program is interface-compatible with the transpiled tests, so that test failures reflect semantic errors rather than signature mismatch. Multiple candidates are sampled per seed and language and executed against the transpiled tests, and only candidates that pass all of them are admitted. 
For Python, the typed reference solution itself joins the source pool. Stage~2 thus establishes execution-verified translation sources in all 25 languages.

\subsubsection{Stage 3: Labeled Data Creation for Reward Model}
\label{sec:data:stage3}

The final stage produces the execution-labeled translation data for reward model training. For each verified source and each of the 24 other target languages, multiple candidate translations are sampled from the policy model and executed against the target-language tests. Candidates that pass all tests are labeled positive, candidates whose pass rate falls below a fixed threshold are labeled negative, and the intermediate band is discarded. 
The stage yields labeled translation samples across the $25{\times}24$ directions.

\subsection{Training Many-to-Many Translation from Execution Feedback}
\label{sec:training}

Training consists of two steps: a reward model is first trained on the labeled data to internalize the execution judge, and policy model optimizes via RL with signals from the reward model.
Rather than using execution outcomes directly as verifiable rewards \citep{lambert2025tulu}, which would require running 25 per-language sandboxes at every RL step, the reward model supplies a graded, execution-grounded score for any candidate in any of the $25{\times}24$ directions (Section \ref{sec:related_work}).

\paragraph{Reward model.}
The reward model $r_\phi$ is a cross-encoder: it jointly encodes a translation prompt $x$ (a source program with its translation direction) and a candidate translation $y$, and returns a scalar score $r_\phi(x,y)$.
A single model serves all directions: it is trained with the Bradley--Terry objective $-\log \sigma\!\left(r_\phi(x,y^{+})-r_\phi(x,y^{-})\right)$ on the pooled Stage-3 pairs, whose negatives are the policy's own execution failures.

\paragraph{Policy optimization.}
RL asks the policy model to translate a program from the Stage-2 verified source pool into one of the other languages, covering $25{\times}24$ directed pairs.
We optimize the policy with a critic-free, group-relative algorithms: GRPO \citep{shao2024deepseekmath}.
For each prompt $x$ the policy samples a group of $G$ candidate translations $\{y_1,\dots,y_G\}$, the reward model scores each as $r_i = r_\phi(x, y_i)$, and each candidate receives its group-relative advantage: $\hat{A}_i = r_i - \mathrm{mean}\{r_1,\dots,r_G\}$,
which weights a clipped policy-gradient update \citep{schulman2017proximal}.
We train each policy in both thinking and non-thinking modes.
To prevent degenerate non-code responses (e.g., refusals to translate) from earning non-trivial scores, any response that lacks a code block or falls below a minimal length is assigned the minimum reward without querying the reward model, so that high reward can be earned only by well-formed translations.

\section{HumanEval-X++}
\label{sec:humanevalxpp}
To fairly evaluate many-to-many translation including the long-tail direction, it is essential that the same problem can be evaluated in a wide variety of target languages.
Existing code translation benchmarks fall into one of two categories: those that target translation among major languages (e.g., HumanEval-X \citep{codegeex}, PolyHumanEval \citep{tao2024_polyHumanEval}, and the MultilingualTrans subset of CodeTransOcean \citep{yan2023codetransocean}), and those that target translation from minor languages into major ones (the NicheTrans subset of CodeTransOcean, which translates from 37 niche languages into 8 popular ones). No existing benchmark starts from a set of major languages and broadens the evaluation to a large number of targets that include minor ones.

To satisfy this requirement in a mechanical and verifiable manner while scaling up the number of languages, we propose HumanEval-X++. This benchmark is an execution-based many-to-many translation benchmark that takes the 6 major languages provided by HumanEval-X \citep{codegeex} as source languages and, using the rule-based transformation of MultiPL-E \citep{multipl-e}, extends the target side to 25 languages.

\subsection{Benchmark Construction}
\label{sec:humanevalxpp:construction}

\begin{figure}[t]
    \centering
    \includegraphics[width=1.0\linewidth]{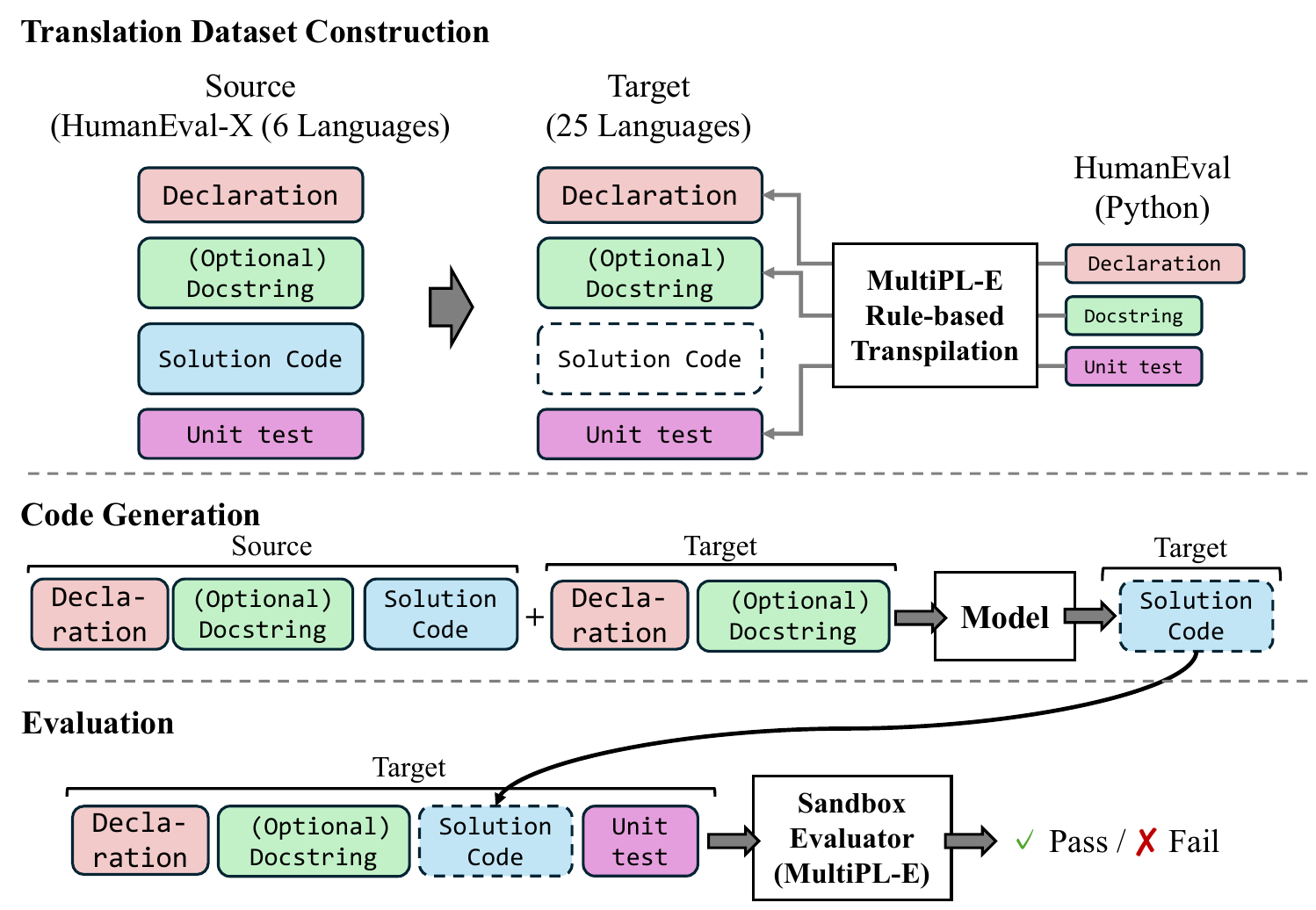}
    \caption{HumanEval-X++ benchmark pipeline. The benchmark is constructed by translating HumanEval problems into multiple target languages via rule-based transpilation. Models generate target-language solutions conditioned on the source declaration and reference solution, along with the target declaration. The generated code is evaluated by executing unit tests in a sandbox environment.}
    \label{fig:humanevalXpp_pipeline}
\end{figure}

The problems covered by HumanEval-X++ take the 164 problems provided by HumanEval~\citep{chen2021_humaneval} as seed problems. Each HumanEval code generation task consists of four components: (1) a declaration (the function name and its arguments), (2) a docstring, (3) a reference function body (the canonical solution), and (4) a unit test comprising multiple test cases that verify functional correctness. Figure~\ref{fig:humanevalXpp_pipeline} illustrates the full pipeline, from seed problems through benchmark construction, generation, and evaluation.

First, as shown in the top of the figure, we use HumanEval-X \citep{codegeex}, which manually extends the Python tasks of HumanEval to 6 major languages, as the source languages. For the target languages, we use the 25 languages supported by the rule-based transpilation of MultiPL-E \citep{multipl-e}, converting the declaration, docstring, and test suite of the Python version of HumanEval into each of these languages. This conversion infrastructure shares the same philosophy of rule-based transpilation and verification by execution as the training data generation pipeline described in Section~\ref{sec:data:stage2} (the Stage~2 export to 25 languages starting from KodCode \citep{kodcode}); the difference is that the starting data is HumanEval rather than KodCode.

Next, in the generation stage shown in the middle of the figure, the translation generation task built on this benchmark takes as model input the source declaration (optionally accompanied by its docstring) together with the reference solution code, followed by the declaration of the target language (again optionally accompanied by its docstring). The model is thereby asked to generate the solution code continuing from the target-language declaration.

Finally, in the evaluation stage shown at the bottom of the figure, the generated target-language solution code is verified by combining the target declaration, the model-generated solution code, and the target test suite, and actually executing them in the MultiPL-E sandbox, yielding an execution-based validation of correctness.
This pipeline guarantees that translation quality of target 25 languages can be evaluated with the same subset of problems. 

\subsection{Language Coverage and Categorization}
\label{sec:humanevalxpp:categorization}

The target languages covered by HumanEval-X++ are, by construction, restricted to those for which MultiPL-E \citep{multipl-e} supports rule-based transformation, amounting to 25 languages in total. In this work, we partition these  languages into three categories --- Popular, Mid-tier, and Long-tail --- based on each language's real-world usage as measured by the GitHub Innovation Graph\footnote{\url{https://github.com/github/innovationgraph/blob/main/data/languages.csv}}.
Concretely, we rank the 25 languages by the number of distinct pushers recorded in the most recent available quarter (2026, Q1), aggregated over all countries; we exclude the aggregate \texttt{EU} entry to avoid double-counting. Whereas CodeTransOcean \citep{yan2023codetransocean} relied on the TIOBE index---a search-popularity proxy over programming languages as a whole---and adopted only a two-way Popular / Niche distinction, we instead adopt a usage-based ranking derived from actual push activity and divide it into three tiers. 

\begin{itemize}
\item \textbf{Popular} (ranks 1-9): JavaScript, Python, TypeScript, Shell, Java, C++, PHP, C\#, Ruby
\item \textbf{Mid-tier} (ranks 10-16): Swift, Go, Rust, Lua, Dart, Perl,R
\item \textbf{Long-tail} (ranks 17-25): Julia, Scala, Haskell, Elixir, Clojure, OCaml, D, Racket, Ada
\end{itemize}


\begin{table}[t]
\centering
\scalebox{0.70}{
 \begin{tabular}{@{}lrrrrrr@{}}
 \toprule
 & \multicolumn{3}{c}{Stage 2} & \multicolumn{3}{c}{Stage 3} \\
 \cmidrule(lr){2-4} \cmidrule(l){5-7}
 Language & Seed & Verified & Rate & Positive & Negative & Pos.\,\% \\
 \midrule
 \multicolumn{7}{@{}l}{\textit{Popular}} \\
 \hspace{4pt}JavaScript & 11,070 & 10,103 & 91.3 & 324,036 & 68,268 & 77.1 \\
 \hspace{4pt}Python     & 11,124 & 11,124 & 100.0 & 327,777 & 35,899 & 86.5 \\
 \hspace{4pt}TypeScript & 10,267 & 8,836 & 86.1 & 267,149 & 110,880 & 66.4 \\
 \hspace{4pt}Shell      & 10,726 & 6,171 & 57.5 & 152,165 & 241,170 & 33.3 \\
 \hspace{4pt}Java       & 10,045 & 7,445 & 74.1 & 204,327 & 188,895 & 49.6 \\
 \hspace{4pt}C++        & 10,981 & 8,305 & 75.6 & 230,752 & 149,340 & 51.6 \\
 \hspace{4pt}PHP        & 11,070 & 9,769 & 88.2 & 280,780 & 103,042 & 67.0 \\
 \hspace{4pt}C\#        & 10,045 & 5,415 & 53.9 & 160,529 & 238,972 & 38.6 \\
 \hspace{4pt}Ruby       & 11,070 & 9,402 & 84.9 & 9,587 & 383,742 & 2.3 \\
 \midrule
 \multicolumn{7}{@{}l}{\textit{Mid-tier}} \\
 \hspace{4pt}Swift & 10,561 & 8,536 & 80.8 & 180,343 & 229,028 & 42.6 \\
 \hspace{4pt}Go    & 9,794 & 6,822 & 69.7 & 183,721 & 205,747 & 45.5 \\
 \hspace{4pt}Rust  & 9,758 & 7,893 & 80.9 & 180,055 & 216,678 & 43.2 \\
 \hspace{4pt}Lua   & 11,070 & 8,537 & 77.1 & 193,833 & 208,519 & 44.8 \\
 \hspace{4pt}Dart  & 9,767 & 7,698 & 78.8 & 180,371 & 199,204 & 44.1 \\
 \hspace{4pt}Perl  & 11,064 & 8,749 & 79.1 & 185,102 & 205,531 & 41.2 \\
 \hspace{4pt}R     & 11,057 & 7,467 & 67.5 & 64,678 & 260,859 & 14.5 \\
 \midrule
 \multicolumn{7}{@{}l}{\textit{Long-tail}} \\
 \hspace{4pt}Julia   & 10,267 & 7,195 & 70.1 & 183,686 & 209,321 & 43.5 \\
 \hspace{4pt}Scala   & 10,753 & 7,285 & 67.7 & 136,005 & 301,623 & 30.2 \\
 \hspace{4pt}Haskell & 9,801 & 4,892 & 49.9 & 50,279 & 357,185 & 12.0 \\
 \hspace{4pt}Elixir  & 11,070 & 7,774 & 70.2 & 65,550 & 372,980 & 14.6 \\
 \hspace{4pt}Clojure & 11,070 & 7,235 & 65.4 & 18,185 & 427,438 & 4.0 \\
 \hspace{4pt}OCaml   & 9,767 & 3,889 & 39.8 & 33,235 & 374,778 & 7.7 \\
 \hspace{4pt}D       & 9,872 & 4,831 & 48.9 & 69,529 & 360,287 & 15.9 \\
 \hspace{4pt}Racket  & 11,070 & 7,657 & 69.2 & 37,279 & 400,124 & 8.3 \\
 \hspace{4pt}Ada     & 10,011 & 1,596 & 15.9 & 15,508 & 434,783 & 3.4 \\
 \midrule
 All & 263,150 & 184,626 & 70.2 & 3,734,461 & 6,284,293 & 34.8 \\
 \bottomrule
 \end{tabular}
}
\caption{Statistics of Stage 2 and 3 at Data Generation.}
\label{tab:stage-2-3-stats}
\end{table}


\begin{table*}[t]
\centering
\scalebox{0.95}{
\small
\begin{adjustbox}{max width=\textwidth}
\begin{tabular}{l cccc|cccc}
\toprule
& \multicolumn{4}{c}{\textbf{Non-Thinking Mode}} & \multicolumn{4}{c}{\textbf{Thinking Mode}} \\
\cmidrule(lr){2-5}\cmidrule(lr){6-9}
& Popular & Mid-tier & Long-tail & All & Popular & Mid-tier & Long-tail & All \\
\midrule
Qwen2.5 Coder 3B Instruct & 75.10 & 45.65 & 28.90 & 49.57 & - & - & - & - \\
Qwen3.5 4B               & 68.08 & 44.53 & 24.19 & 45.08 & 64.93 & 46.67 & 24.19 & 44.59 \\
Qwen3.5 4B-GRPO          & \textbf{81.94} & \textbf{66.07} & \textbf{30.52} & \textbf{58.29} & \textbf{79.07} & \textbf{59.13} & \textbf{32.83} & \textbf{56.18} \\
\midrule
Qwen2.5 Coder 7B Instruct & 78.08 & 54.68 & 34.94 & 55.40 & - & - & - & - \\
Qwen3.5 9B               & 75.19 & 51.69 & 31.84 & 52.42 & 70.84 & 54.27 & 33.60 & 52.31 \\
Qwen3.5 9B-GRPO          & \textbf{85.57} & \textbf{72.24} & \textbf{43.30} & \textbf{66.04} & \textbf{82.31} & \textbf{63.57} & \textbf{42.18} & \textbf{62.07} \\
\bottomrule
\end{tabular}
\end{adjustbox}
}
\caption{Average pass@1 accuracy (\%) across Popular / Mid-tier / Long-tail target languages.}  
\label{tab:humanevalXpp_result}
\end{table*}

\begin{table}[t]
\centering
\scalebox{0.95}{
\begin{tabular}{@{}lcc@{}}
\toprule
Model & \makecell{Non-Thinking} & \makecell{Thinking} \\
\midrule
Qwen3.5-4B & 14.58 & 11.74 \\
Qwen3.5-4B + GRPO & \textbf{15.52} & \textbf{17.30} \\
\midrule
Qwen3.5-9B & 26.76 & 23.08 \\
Qwen3.5-9B + GRPO & \textbf{28.78} & \textbf{29.70} \\
\bottomrule
\end{tabular}
}
\caption{CodeScope executable pass@1 accuracy (\%) over 5,317 translations. Maximum generation length is 8,192.}
\label{tab:codescope_result}
\end{table}

\begin{table*}[t]
\centering
\begin{adjustbox}{max width=\textwidth}
\newcommand{\HM}[1]{#1}
\newcommand{\NAcell}{n/a}
\begin{tabular}{l*{29}{c}}
\toprule
& \multicolumn{29}{c}{\textbf{Target Language}} \\
& \multicolumn{10}{c}{\textbf{Popular}} & \multicolumn{8}{c}{\textbf{Mid-tier}} & \multicolumn{10}{c}{\textbf{Long-tail}} & \\
\cmidrule(lr){2-11}\cmidrule(lr){12-19}\cmidrule(lr){20-29}
& JS & Python & TS & Shell & Java & C++ & PHP & C\# & Ruby & (AVG)
& Swift & Go & Rust & Lua & Dart & Perl & R & (AVG)
& Julia & Scala & Haskell & Elixir & Clojure & OCaml & D & Racket & Ada & (AVG)
& (All-AVG) \\
\midrule
\multicolumn{30}{l}{\textit{Qwen3.5 9B}} \\
cpp  & \NAcell    & \HM{85.6} & \HM{89.4} & \HM{76.6} & \HM{54.1} & \HM{84.7} & \HM{83.1} & \HM{87.3} & \HM{28.1} & \HM{73.6}
     & \HM{71.9} & \HM{69.3} & \HM{75.5} & \HM{26.9} & \HM{84.0} & \HM{9.4}  & \HM{39.4} & \HM{53.7}
     & \HM{36.7} & \HM{73.6} & \HM{34.8} & \HM{16.3} & \HM{26.9} & \HM{38.3} & \HM{21.3} & \HM{31.3} & \HM{13.5} & \HM{32.5}
     & \HM{52.4} \\
go   & \HM{91.9} & \HM{93.8} & \HM{84.8} & \HM{55.4} & \HM{79.6} & \HM{81.3} & \HM{86.3} & \HM{83.4} & \HM{25.6} & \HM{75.8}
     & \HM{53.1} & \NAcell    & \HM{72.3} & \HM{28.8} & \HM{74.4} & \HM{10.6} & \HM{51.9} & \HM{48.5}
     & \HM{39.2} & \HM{54.7} & \HM{34.2} & \HM{23.1} & \HM{35.0} & \HM{27.9} & \HM{22.6} & \HM{30.0} & \HM{11.5} & \HM{30.9}
     & \HM{52.1} \\
java & \HM{92.5} & \HM{95.6} & \HM{86.1} & \HM{60.5} & \NAcell    & \HM{87.5} & \HM{83.1} & \HM{90.4} & \HM{24.4} & \HM{77.5}
     & \HM{63.8} & \HM{60.8} & \HM{79.4} & \HM{20.6} & \HM{79.5} & \HM{8.1}  & \HM{47.5} & \HM{51.4}
     & \HM{41.1} & \HM{65.4} & \HM{34.8} & \HM{23.1} & \HM{38.1} & \HM{31.8} & \HM{18.7} & \HM{30.6} & \HM{15.4} & \HM{33.2}
     & \HM{53.3} \\
js   & \NAcell    & \HM{93.1} & \HM{75.9} & \HM{59.9} & \HM{84.1} & \HM{85.0} & \HM{81.9} & \HM{89.2} & \HM{32.5} & \HM{75.2}
     & \HM{60.0} & \HM{73.2} & \HM{72.3} & \HM{28.1} & \HM{80.8} & \HM{6.9}  & \HM{46.9} & \HM{52.6}
     & \HM{43.0} & \HM{51.6} & \HM{29.0} & \HM{17.5} & \HM{32.5} & \HM{35.1} & \HM{20.6} & \HM{25.0} & \HM{17.3} & \HM{30.2}
     & \HM{51.7} \\
py   & \HM{88.8} & \NAcell    & \HM{90.5} & \HM{51.0} & \HM{76.4} & \HM{84.4} & \HM{81.9} & \HM{89.2} & \HM{28.8} & \HM{73.9}
     & \HM{58.8} & \HM{63.4} & \HM{78.1} & \HM{26.9} & \HM{76.3} & \HM{8.1}  & \HM{42.5} & \HM{50.6}
     & \HM{40.5} & \HM{65.4} & \HM{36.1} & \HM{25.6} & \HM{36.9} & \HM{31.2} & \HM{16.8} & \HM{31.9} & \HM{14.7} & \HM{33.2}
     & \HM{51.8} \\
rust & \HM{82.5} & \HM{88.1} & \HM{77.8} & \HM{54.1} & \HM{79.6} & \HM{81.9} & \HM{81.9} & \HM{90.4} & \HM{40.0} & \HM{75.2}
     & \HM{59.4} & \HM{62.7} & \HM{80.6} & \HM{28.8} & \HM{78.8} & \HM{11.9} & \HM{51.3} & \HM{53.4}
     & \HM{42.4} & \HM{67.3} & \HM{29.7} & \HM{23.8} & \HM{30.6} & \HM{27.9} & \HM{16.8} & \HM{28.1} & \HM{12.2} & \HM{31.0}
     & \HM{53.1} \\
\cmidrule(lr){1-30}
(AVG) & \HM{88.3} & \HM{92.0} & \HM{82.0} & \HM{55.8} & \HM{80.9} & \HM{84.0} & \HM{83.0} & \HM{88.3} & \HM{29.9} & \HM{75.2} & \HM{61.1} & \HM{65.9} & \HM{76.3} & \HM{26.7} & \HM{79.0} & \HM{9.2} & \HM{46.6} & \HM{51.7} & \HM{40.5} & \HM{63.0} & \HM{33.1} & \HM{21.6} & \HM{33.3} & \HM{32.0} & \HM{19.5} & \HM{29.5} & \HM{14.1} & \HM{31.8} & \HM{52.4} \\
\midrule
\multicolumn{30}{l}{\textit{Qwen3.5 9B-GRPO}} \\
cpp  & \NAcell    & \HM{90.0} & \HM{91.3} & \HM{93.0} & \HM{59.9} & \HM{89.8} & \HM{88.1} & \HM{95.5} & \HM{86.9} & \HM{86.8}
     & \HM{68.1} & \HM{75.8} & \HM{76.8} & \HM{83.1} & \HM{87.8} & \HM{66.9} & \HM{53.8} & \HM{73.2}
     & \HM{77.8} & \HM{74.2} & \HM{43.2} & \HM{24.4} & \HM{35.0} & \HM{41.6} & \HM{38.7} & \HM{23.1} & \HM{16.0} & \HM{41.6}
     & \HM{65.9} \\
go   & \HM{93.1} & \HM{94.4} & \HM{93.0} & \HM{54.1} & \HM{79.0} & \HM{86.3} & \HM{90.0} & \HM{91.7} & \HM{90.6} & \HM{85.8}
     & \HM{70.0} & \NAcell    & \HM{74.8} & \HM{76.3} & \HM{85.9} & \HM{71.9} & \HM{57.5} & \HM{72.7}
     & \HM{81.6} & \HM{66.7} & \HM{37.4} & \HM{37.5} & \HM{37.5} & \HM{28.6} & \HM{52.3} & \HM{24.4} & \HM{14.7} & \HM{42.3}
     & \HM{66.2} \\
java & \HM{92.5} & \HM{95.0} & \HM{91.8} & \HM{62.4} & \NAcell    & \HM{88.1} & \HM{92.5} & \HM{94.3} & \HM{89.4} & \HM{88.2}
     & \HM{76.9} & \HM{70.6} & \HM{80.0} & \HM{77.5} & \HM{84.0} & \HM{68.8} & \HM{56.9} & \HM{73.5}
     & \HM{83.5} & \HM{73.6} & \HM{43.9} & \HM{35.0} & \HM{41.3} & \HM{35.1} & \HM{57.4} & \HM{28.1} & \HM{18.6} & \HM{46.3}
     & \HM{68.2} \\
js   & \NAcell    & \HM{93.8} & \HM{93.0} & \HM{63.1} & \HM{86.6} & \HM{87.5} & \HM{88.8} & \HM{93.0} & \HM{85.0} & \HM{86.3}
     & \HM{73.1} & \HM{76.5} & \HM{74.8} & \HM{71.9} & \HM{84.6} & \HM{71.3} & \HM{53.8} & \HM{72.3}
     & \HM{77.2} & \HM{68.6} & \HM{40.0} & \HM{31.9} & \HM{34.4} & \HM{37.0} & \HM{50.3} & \HM{22.5} & \HM{16.0} & \HM{42.0}
     & \HM{65.6} \\
py   & \HM{87.5} & \NAcell    & \HM{85.4} & \HM{56.7} & \HM{79.6} & \HM{88.8} & \HM{86.9} & \HM{92.4} & \HM{85.0} & \HM{82.8}
     & \HM{71.9} & \HM{67.3} & \HM{74.2} & \HM{71.3} & \HM{82.7} & \HM{65.6} & \HM{53.8} & \HM{69.5}
     & \HM{73.4} & \HM{71.7} & \HM{40.0} & \HM{33.8} & \HM{37.5} & \HM{36.4} & \HM{52.9} & \HM{31.3} & \HM{14.1} & \HM{43.4}
     & \HM{64.2} \\
rust & \HM{86.3} & \HM{88.8} & \HM{86.7} & \HM{53.5} & \HM{83.4} & \HM{87.5} & \HM{85.6} & \HM{95.5} & \HM{83.8} & \HM{83.5}
     & \HM{66.9} & \HM{75.8} & \HM{83.9} & \HM{78.1} & \HM{84.0} & \HM{62.5} & \HM{54.4} & \HM{72.2}
     & \HM{79.7} & \HM{79.2} & \HM{39.4} & \HM{36.3} & \HM{38.8} & \HM{31.8} & \HM{45.2} & \HM{32.5} & \HM{15.4} & \HM{44.2}
     & \HM{66.2} \\
\cmidrule(lr){1-30}
(AVG) & \HM{89.9} & \HM{92.6} & \HM{90.5} & \HM{58.3} & \HM{83.7} & \HM{87.6} & \HM{88.6} & \HM{93.7} & \HM{86.8} & \HM{85.6} & \HM{71.1} & \HM{73.2} & \HM{77.4} & \HM{76.4} & \HM{84.8} & \HM{67.8} & \HM{55.0} & \HM{72.2} & \HM{78.9} & \HM{72.3} & \HM{40.6} & \HM{33.1} & \HM{37.4} & \HM{35.1} & \HM{49.5} & \HM{27.0} & \HM{15.8} & \HM{43.3} & \HM{66.0} \\
\bottomrule
\end{tabular}
\end{adjustbox}
\caption{Per-language translation accuracy (pass@1, \%) for Qwen3.5\_9B and its GRPO-trained variant.}  
\label{tab:heatmap_bycategory}
\end{table*}


\section{Experiment}
\label{sec:experiment}

\subsection{Settings}

\subsubsection{Data Generation}
Stage~1 ramdomly sampled 10,000 seed python code problems from KodCode(-V1), resulting in 11,124 programs (9,127 distinct problems) with 11.4 verified assertions per program on average after test case synthesis, type annotation, and filtering by sandboxed execution. 
We request the LLM to synthesize the assertions five times per target program with one prompt requesting at least five test cases. This yields roughly 25 raw candidates per target in the nominal case.
The threshold of the assertion filtering is that each assertion must execute with exit code 0 and status OK, and at least 5 valid assertions must remain per program. 
The data-generation LLM used for synthesizing tests in Stage~1 is Qwen3-Coder-Next (FP8). 
Its sampling temperature is set as 1.0, top-p 0.95, top-k 40, up to 2,048 new tokens, with 5 samples per prompt. 

Stage~2 expands the seeds to all 25 languages, producing 184,626 execution-verified source programs for each programs.
Base LLM used in the Stage~2 is Qwen3-Coder-Next (FP8).
Stage~3 rollouts translations using base model and attach success/failure labels by execution, yielding 3,734,461 positive and 6,284,293 negative samples out of 10,724,077 rollouts across the four model conditions. It yields 717,083 labeled preference pairs over the $25{\times}24$ directions. 
The base model used in Stage~3 is Qwen3.5-4B / Qwen3.5-9B, which is also the checkpoint that RL later optimizes.
Its sampling temperature is set as 1.0, top-p 0.95, top-k 20, presence penalty 1.5, up to 4,096 new tokens in non-thinking mode and 12,288 in thinking mode, with 5 samples per direction.

\subsubsection{Training}
In reward model training, the model is initialized from Qwen3.5-35B-A3B and trained on the Stage~3 pair data described in Section~\ref{sec:training} using LoRA (rank 16, alpha 32, applied to all linear layers). 
The model is trained for 2,774 steps (1 epoch), batch size 256, learning rate 1e-5, AdamW optimizer, cosine scheduler with ~139 (5\% of total) warmup steps.
In reinforcement learning, the model is initialized from Qwen3.5-4B / Qwen3.5-9B and trained on the source program described in Section~\ref{sec:training}.
The model optimizes the policy against the trained reward model, sampling a group of 4 candidates per prompt.
During rollout, we sample 4 responses per prompt with temperature 1.0, top-p 1.0, and top-k disabled, with a maximum prompt length of 1,024 tokens and a maximum response length of 4,096 tokens. These settings follow the default rollout configuration provided by verl.
We train each policy model in both thinking and non-thinking modes.
The model is trained for 100 steps (1 epoch), batch size 256, learning rate 1e-6, AdamW optimizer, constant scheduler with 0 warmup steps.


\subsubsection{Evaluation}
Our experiment uses HumanEval-X++ as our main benchmark (Section~\ref{sec:humanevalxpp}). For each language pair, we evaluate all samples for which MultiPL-E's rule-based transformation was applied successfully (up to a maximum of 164 per language pair). To assess translation ability in a stringent setting, we remove the docstring from both the source and the target, so that the model must translate purely from code without any natural-language description.
We measure the correctness of the generated target-language code using the pass@\(k\) metric introduced in HumanEval~\citep{chen2021_humaneval}; in our experiments, we generate a single completion per problem and thus report pass@1,  i.e., standard accuracy.

We also evaluate on CodeScope~\citep{yan-etal-2024-codescope}. It contains 5,317 program-translation examples over 182 source--target directions among 14 languages: C, C\#, C++, D, Delphi, Go, Java, JavaScript, Kotlin, PHP, Perl, Python, Ruby, and Rust.
All directions ($14 \times 13$ = 182) are evaluated for all the languages.
Evaluated sample size is near-uniform per direction: 24--30 examples per direction (30 for 130 of the 182 directions), which gives 312--390 examples per source language and 379--385 per target language.
The evaluation approach of CodeScope differs from HumanEval-X++. While HumanEval-X++ extends function-level HumanEval-X tasks to 25 target languages by rule-based transformation, CodeScope uses independent programs with input--output tests.
The official CodeScope code checks only one input--output pair for each generated program. Because each task provides multiple pairs, we judge a translation as correct only if it passes all pairs. We first run the reference solutions and remove any task with an inconsistent pair. This reduces the set from 5,382 to 5,317 examples.


\subsubsection{Other Settings}
All experiments were conducted on multiple GH200 nodes with 90 GB of GPU memory per device. We used the model parameters released on the HuggingFace. Due to the limited computational resources available, we performed only a single hyperparameter sweep for the pipeline. Consequently, all experimental results are based on a single run and are reported without statistical measures.

\subsection{Main Result}

Table~\ref{tab:humanevalXpp_result} reports the results on  HumanEval-X++.
Our method improves over the base model in all evaluated conditions
(two model scales $\times$ two inference modes).
In the best configuration under Non-Thinking mode, the 4B model improves from 45.08 to 58.29 ($+13.21$)
and the 9B model from 52.42 to 66.04 ($+13.62$) on average across all the languages; Comparable gains hold under Thinking mode
(4B: 44.59 $\to$ 59.15, 9B: 52.31 $\to$ 63.11).
Broken down by tier (Non-Thinking, GRPO), the largest improvement falls on Mid-tier languages,
reaching $+21.54$ for 4B and $+20.55$ for 9B.
By contrast, the Long-tail gain for 4B is only $+6.33$, smaller than the Popular gain ($+13.86$).
In other words, there is no monotone relation in which the more long-tail the target language, the larger the gain.
We further analyze this by language in Section~\ref{sec:result_by_lang}.



On CodeScope, our method also improves the performance on both models, with the strongest result of 29.70 for the 9B model in Thinking mode (Table~\ref{tab:codescope_result}).

\subsection{Result by Language}
\label{sec:result_by_lang}
To further analyze the result, we decompose a representative scenario (9B, Non-Thinking, base vs.\ GRPO) by target language (Table~\ref{tab:heatmap_bycategory}).
The largest gains accrue to languages on which the base model scored extremely poorly: Perl 9.2$\to$67.8, Ruby 29.9$\to$86.8, Lua 26.7$\to$76.4, Julia 40.5$\to$78.9, D 19.5$\to$49.5.
These languages belong to no single tier: Ruby is Popular, Perl and Lua are Mid-tier, and Julia and D are Long-tail.
From this result, we infer that the large average Mid-tier gain is not a property of the tier itself, but two big recovery (Perl and Lua) falling into it.

We examined the nature of this recovery through failure analysis, concluding that the largest contribution to the gain is the reduction of code that never reaches execution.
For Qwen3.5\_9B and its GRPO variant, we analyze the $22{,}755$ translations from the six source languages into the 25 target that both models were evaluated on in common.
The two models succeed on $11{,}886$ and $15{,}015$ respectively,
an increase of $3,129$ cases.
Of these, compilation errors fall from $4,222$ to $2,538$ cases,
a reduction of $1,684$ cases that accounts for more than half of the increase.

However, some languages within Long-tail do not recover: Gains are small for Ada (14.1$\to$15.8\%), OCaml (32.0$\to$35.1\%), Clojure (33.3$\to$37.4\%),
and Racket (29.5 $\to$ 27.0\%).
These are functional and declarative languages that are syntactically distant from the source languages,
suggesting causes such as an insufficient knowledge of the language itself
and a failure to conform to the required output format.

\subsection{Qualitative Analysis}
\label{sec:qualitative}

\begin{figure}[t]
\centering
\includegraphics[width=0.99\columnwidth]{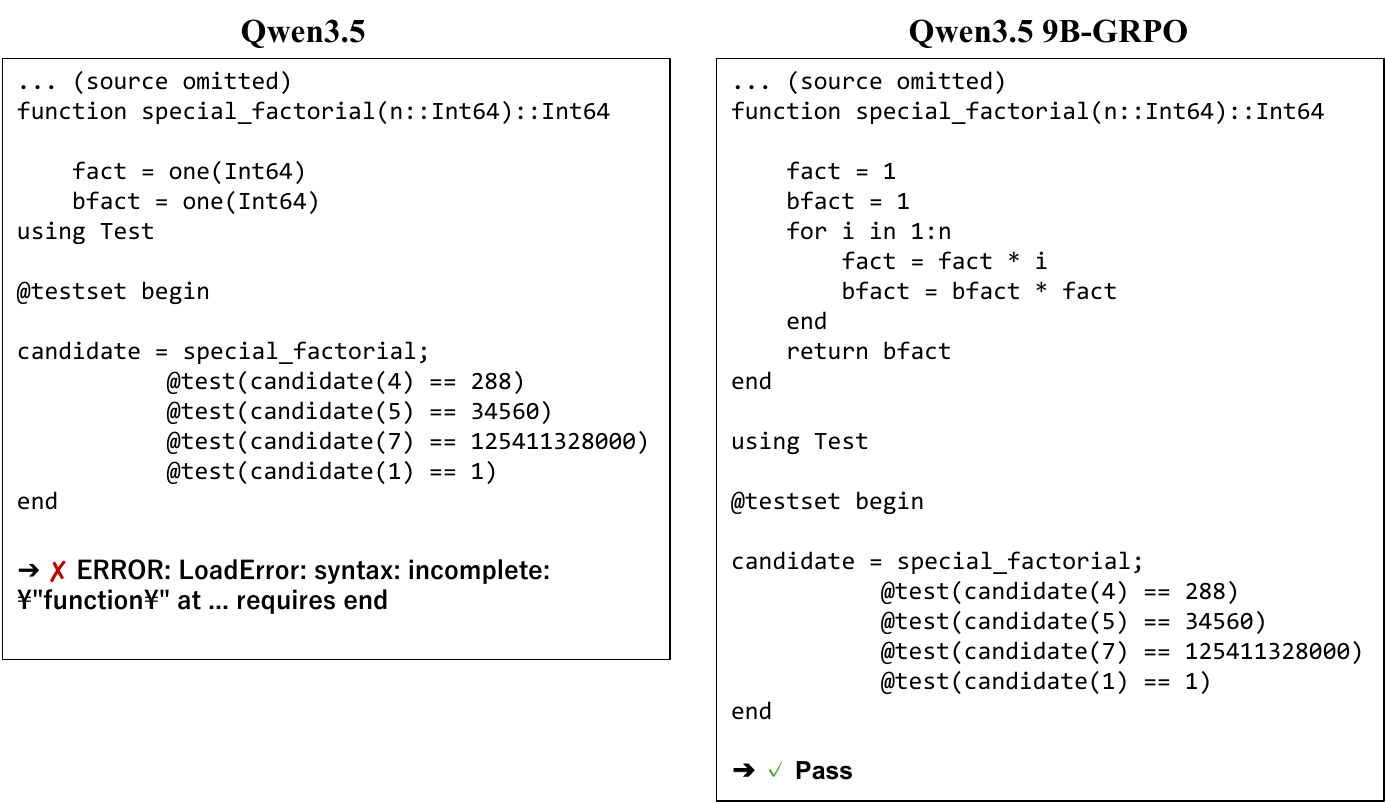}
\caption{An example of C++ to Julia translation. Without RL, the output fails to compile
because the body of the solution is missing and the \texttt{end} closing the function is omitted.
After GRPO, the model emits the same algorithm in conformance with Julia's conventions and passes the tests.}
\label{fig:case_cpp2jl}
\end{figure}

We illustrate a recovery example with a concrete case, translation from C++ to Julia.
Over the 158 evaluated samples of this translation task, compilation errors drop from 62.7\% (99 cases)
for Qwen3.5 9B to 22.2\% (35 cases) after GRPO,
which closely matches the rise in the number of samples passing the tests,
from 36.7\% (58 cases) to 77.8\% (123 cases).
In Figure~\ref{fig:case_cpp2jl}, Qwen3.5 9B outputs the body of the solution code
and further omits the \texttt{end} closing the function, so it is rejected at compile time.

\section{Conclusion}
We proposed NicheCodeTranslator, which expands verifiable Python seeds into an execution-validated multilingual pool, trains a reward model on execution-labeled preferences, and optimizes the policy with GRPO using this learned reward instead of per-language sandboxes, scaling stably to 600 directions across 25 languages. We also introduced HumanEval-X++, an execution-based benchmark extending HumanEval-X to 25 target languages. Our method consistently improves over baselines on HumanEval-X++ and CodeScope, paving the way toward further bootstrapping the quality of many-to-many translation for programming languages.



\bibliography{references}

\end{document}